\def\year{2019}\relax
\documentclass[letterpaper]{article} 
\usepackage{aaai19}  
\usepackage{times}  
\usepackage{helvet}  
\usepackage{courier}  
\usepackage{url}  
\usepackage{graphicx}  
\usepackage{subfigure}
\usepackage{amsmath,amssymb} 
\usepackage{algorithm}
\usepackage{algorithmicx}
\usepackage{algpseudocode}
\usepackage{verbatim}
\usepackage{tabularx}
\usepackage{booktabs} 
\begin{document}
%
\title{DeepIR: A Deep Semantics Driven Framework for Image Retargeting}
\author{Jianxin Lin\thanks{The first two authors contributed equally to this work}, Tiankuang Zhou\footnotemark[1], Zhibo Chen\\
University of Science and Technology of China, Hefei, China\\
\{linjx, zhoutk\}@mail.ustc.edu.cn, chenzhibo@ustc.edu.cn\\
}
\maketitle
\newcommand{\citelp}[1]{\citeauthor{#1}~\shortcite{#1}}
\begin{abstract}
We present \emph{Deep Image Retargeting} (\emph{DeepIR}), a coarse-to-fine framework for content-aware image retargeting. Our framework first constructs the semantic structure of input image with a deep convolutional neural network. Then a uniform re-sampling that suits for semantic structure preserving is devised to resize feature maps to target aspect ratio at each feature layer. The final retargeting result is generated by coarse-to-fine nearest neighbor field search and step-by-step nearest neighbor field fusion. We empirically demonstrate the effectiveness of our model with both  qualitative and quantitative results on widely used RetargetMe dataset.
\end{abstract}

\section{Introduction}
The heterogeneity of the display devices have imposed indispensable demand for appropriately adapting image into screens with different resolutions. The image retargeting techniques meanwhile have been proposed to resize images to arbitrary sizes while keeping the important content of original images. These content-aware image retargeting methods, such as seam carving \cite{avidan2007seam}, multi-operator \cite{rubinstein2009multi,zhu2016saliency}, and streaming video  \cite{krahenbuhl2009system}, try to resize the image to target resolution by preserving the important content information and shrinking the unimportant regions of the original image.

Traditional content-aware methods usually utilize one saliency map to define the significance of each pixel. However, saliency detection is designed from attention mechanism, one saliency map has its limitations not only on representing the high-level semantic content, but also on integrating both high-level semantic content and low-level details. In recent years, Convolutional Neural Network (CNN) has shown its superior performance in many high-level computer vision problems \cite{he2016deep,long2015fully} and low-level image processing problems \cite{srgan,aaai2018ours,rednet}. One main advantage of CNN is its better ability to extract multi-level semantic information and better representations for semantic structures \cite{zeiler2014visualizing}. Therefore, how to utilize CNN to guide image retargeting is interesting and important to explore.

\begin{figure}[t]
\centering
\includegraphics[width=8.0cm]{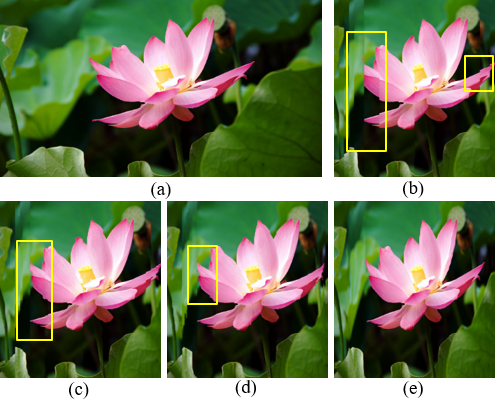}
	\caption{Illustration of DeepIR on image retargeting. (a) Original image. (b)-(e) Coarse-to-fine image retargeting refinement. (e) Final retargeted image. At the coarsest layer, the retargeted image (b) maintains the main semantic structure while suffering unsmoothness. The coarse-to-fine refinement preserves the main semantic structure and content smoothness.}
	\label{fig:coarse2fine}
\end{figure}

In this paper, we propose a new framework for content-aware image retargeting. Instead of applying image retargeting techniques on image pixel level directly, our approach uses a pre-trained deep CNN, such as VGG-19 \cite{simonyan2014very}, to construct a feature space in which image retargeting is performed. In order to resize original image to the target aspect ratios in the deep feature space, we devise a uniform re-sampling (UrS) that uniformly removes columns/rows in a cumulative columns/rows obscurity map. Such a UrS ensures the semantic structure completeness of resized feature maps, and content smoothness is also preserved in the final retargeted image as illustrated in Figure \ref{fig:coarse2fine}(e). Then a coarse-to-fine nearest neighbor field (NNF) search \cite{Liao:2017:VAT:3072959.3073683} is used to find spatial correspondence between intermediate feature layers of original image and retargeted image. At each layer, two NNFs obtained from reconstructed features and retargeted features respectively are fused to achieve a combination of high level and low level information, which is called as step-by-step NNF fusion.

The main contribution of our work includes three aspects:
\begin{itemize}
  \item We propose a new Deep Image Retargeting (DeepIR) framework that retargets images in the deep feature space. We demonstrate that DeepIR is effective for semantic content and visual quality preservation.
  \item We propose a UrS that suits for retargeting in the deep feature space.
  \item We propose a step-by-step NNF fusion that effectively combines the high level semantic content and low level details.
\end{itemize}

The remaining parts are organized as follows. We introduce related work in Section \ref{related work} and present the details of our method in Section \ref{framework}. Then we report experimental results in Section \ref{experiment} and conclude in Section \ref{conclusion}.

\section{Related Works}\label{related work}
Numerous works have been carried out for image retargeting in the past decades. Unlike traditional image retargeting methods, such as uniform Scaling (SCL), recent developments in image retargeting usually seek to change the size of the image while maintaining the important content. By using face detectors \cite{viola2001rapid} or the visual saliency detection methods \cite{itti1998model} to detect important regions in the image, one simple way to resize image is using Cropping (CR) to eliminate the unimportant region from the image. However, directly eliminating region by CR may result in information loss. Seam Carving (SC) \cite{avidan2007seam} is proposed to iteratively remove an $8$-connected seam in the image to preserve the visual saliency. To avoid drawbacks of using single retargeting method, multi-operator techniques (MULTIOP) \cite{rubinstein2009multi,zhu2016saliency} combine SC, SCL and CR to resize the image based on the defined optimal energy cost, such as image retargeting quality assessment metrics. \citelp{pritch2009shift} described a Shift-Map (SM) technique to remove or add band regions instead of scaling or stretching images. All the above methods resize the image by removing discrete regions. Other approaches also put attempts on resizing the image in continuous and summarization perspectives.

Continuous retargeting methods continuously transform image to the target size and have been realized through image warping or the mapping by constraining deformation and smoothness \cite{wang2008optimized,wolf2007non,krahenbuhl2009system,guo2009image,lin2014object}. \citelp{wang2008optimized} presented a ``scale-and-stretch'' (SNS) warping method that operates by iteratively computing optimal local scaling factors for each local region and updating a warped image that matches these scaling factors as close as possible. \citelp{wolf2007non} described a nonhomogeneous warping (WARP) for video retargeting, where a transformation that respects the analysis shrinks less important regions more than important ones. \citelp{krahenbuhl2009system} presented a simple and interactive framework called Streaming Video (SV) that combines key frame based constraint editing with numerous automatic algorithms for video analysis. The key component of their framework is a non-uniform and pixel accurate warping considering automatic as well as interactively defined features. \citelp{guo2009image} constructed a saliency-based mesh representation for an image, which enables preserving image structures during retargeting. To avoid object deformation caused by warping, \citelp{lin2014object} utilized the information of matched objects rather than that of matched pixels during warping process, which allows the generation of an object significance map and the consistent preservation of objects. In \cite{karni2009energy}, authors proposed a energy minimization based shape deformation method (LG) in which the set of ``legal transformations'' being not considered to be distorted is expressed.

Summarization based retargeting methods \cite{simakov2008summarizing,cho2008patch,barnes2009patchmatch,wu2010resizing} resize image by eliminating insignificant patches and maintaining the coherence between original and retargeted image. \citelp{simakov2008summarizing} measured the bidirectional patch similarity (i.e., completeness and coherence) between two images and iteratively change the original image's size to retargeted image's. \citelp{cho2008patch} breaked an image to non-overlapping patches and retargeted image is constructed from the patches with ``patch domain'' constrains. \citelp{barnes2009patchmatch} proposed a fast randomized algorithm called PatchMatch to find dense NNF for patches between two images, and retargeted image can be obtained by the similar retargeting method in \cite{simakov2008summarizing}. \citelp{wu2010resizing} analyzed the ``translational symmetry'' widely existed in the real-world images, and summarize the image content based on symmetric lattices.

Most previous content-aware image retargeting algorithms leverage saliency detection or object detection information to select important regions. However, real-world images usually contain complex and abundant semantic information that should be modeled in a more appropriate manner. Recently, CNN has shown its superior performance in many high-level computer vision problems \cite{he2016deep,long2015fully} and low-level image processing problems \cite{srgan,aaai2018ours,rednet}. \citelp{cho2017weakly} first applied deep CNN to image retargeting. A weakly- and self-supervised deep convolutional neural network (WSSDCNN) is trained for shift map prediction. However, quantitative results of WSSDCNN only indicate its superiority over SCL or SC. Compared with WSSDCNN, our method doesn't need any training procedure when a pre-trained CNN is given. In addition, our DeepIR shows comparable performance against SOTA methods (i.e., MULTIOP, SV) in Section \ref{experiment}.

\section{Approach}\label{framework}
\begin{figure}
\centering
\includegraphics[width=8.0cm]{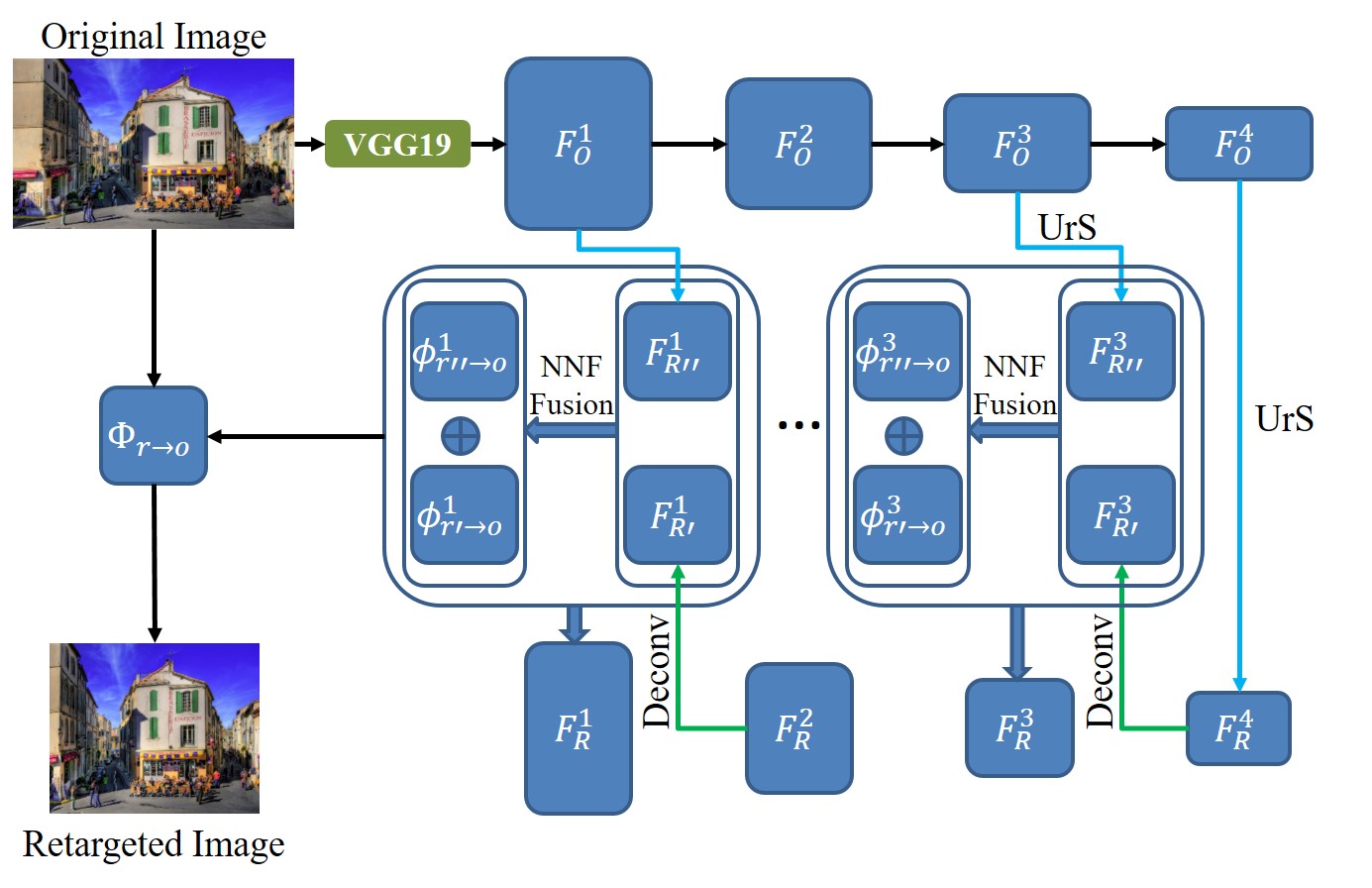}
	\caption{Deep image retargeting (DeepIR) framework.}
	\label{fig:deepIRframwork}
\end{figure}
Figure \ref{fig:deepIRframwork} shows the overall architecture of the proposed DeepIR framework. Our model mainly includes three parts: deep feature construction, deep feature retargeting and retargeted image reconstruction. The details will be presented in the following sections.

\subsection{Pre-Processing}
In order to obtain deep feature space of original image, we utilize the VGG-19 network \cite{simonyan2014very} that is pre-trained on the ImageNet database \cite{russakovsky2015imagenet} as our deep CNN. For original image $O$, we can obtain a pyramid of its feature maps as $\{F_O^L\}$ $(L=1...4)$. The reason we choose the first four layers of VGG-19 network as our feature space is that the resolutions of feature maps in the higher layers are too small and it will be too difficult to reconstruct retargeted image based on them. The feature maps of original image are 3D tensors with $h_O^L \times w_O^L \times c_O^L$ for each layer, where $h_O^L$ is height, $w_O^L$ is width and $c_O^L$ is channel number for each $F_O^L$.

\subsection{Uniform Re-Sampling}
\begin{figure}
\centering
\includegraphics[width=8.0cm]{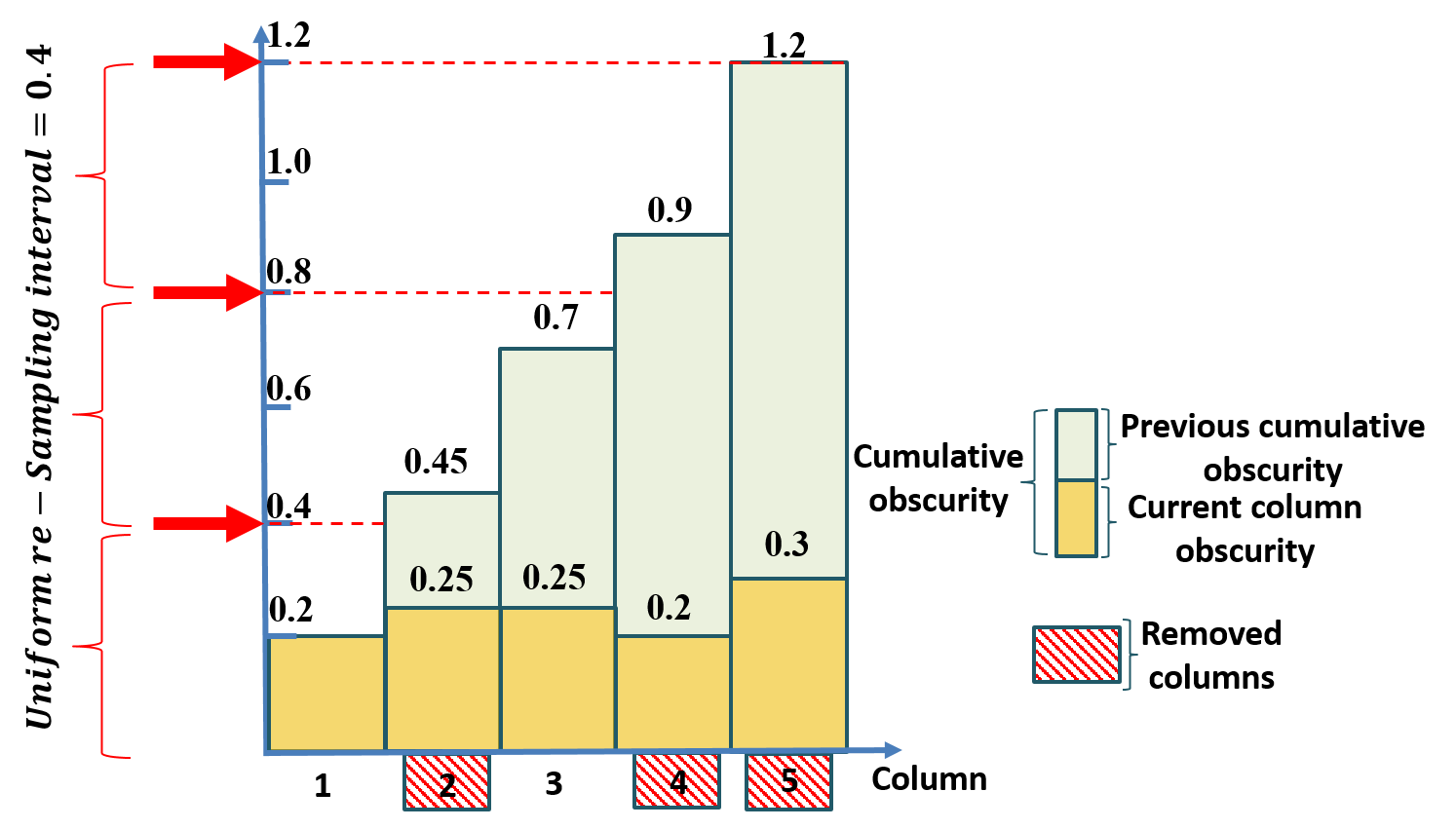}
	\caption{ An example of UrS method resizing $5$ columns to $2$ columns. For each column other than the first one, there are two overlapping bins, in which the lower one represents the normalized obscurity and the higher one represents the cumulative obscurity.
Then a uniform sampling is applied on the cumulative obscurity map. Obscurity sum $s(5) = 1.2$. Sampling interval $\tau = s(5)/(5-2)=0.4$. The bins marked by red tilted lines represent the removed columns.}
	\label{fig:SmoothColumnRemoval}
\end{figure}

In order to resize original image to the target aspect ratios in the deep feature space, we devise a uniform re-sampling (UrS) that uniformly removes columns/rows in a cumulative columns/rows obscurity map. As \cite{zeiler2014visualizing} proved that region of higher semantic significance results in stronger activation in feature maps, given feature maps $F_O^L$, an importance map is first computed as:
\begin{equation}
m_O^L(i,j)=\sum_{c=1}^{c_O^L}F_O^L(i,j,c),
\label{eq:energyf}
\end{equation}
where $i,j,c$ are the height, width and channel index respectively. Then without loss of generality, the obscurity map, which gives higher weight to less important (obscure) columns, is computed as:
\begin{equation}
u_O^{L,w}(j)=-\sum_{i=1}^{h_O^L}m_O^L(i,j).
\label{eq:prob_column1}
\end{equation}
Then the obscurity map $u_O^{L,w}$ is normalized by min-max normalization, expressed as $\hat{u}_O^{L,w}$. After that, the cumulative obscurity map is obtained as:
\begin{equation}
s_O^{L,w}(j)=
\left\{\begin{matrix}
 & \hat{u}_O^{L,w}(1),  \quad j=1,\\
 & s_O^{L,w}(j-1)+\hat{u}_O^{L,w}(j),\quad  j>1.
\end{matrix}\right.
\label{eq:prob_column2}
\end{equation}
Given a retargeting aspect ratio $\epsilon$, $(w_O^{L}-\epsilon w_O^{L})$ columns are to be removed. Then a uniform sampling is performed on the $s_O^{L,w}$ with sampling interval $\tau=s_O^{L,w}(w_O^L)/(w_O^L -\epsilon w_O^L)$. And obscure columns $\mathcal{R}$ to be removed are defined as below:
\begin{equation}
\mathcal{R}=\{j|s_O^{L,w}(j-1) \leq r*\tau < s_O^{L,w}(j),\exists r \in \{1,...,w_O^L -\epsilon w_O^L\} \}.
\label{eq:prob_column1}
\end{equation}
Denoting all original columns as $\mathcal{A}$, the preserved columns are $\mathcal{P} =\mathcal{A} -\mathcal{R} $. We call this process as uniform re-sampling that is illustrated in Figure \ref{fig:SmoothColumnRemoval}. The final retargeted feature maps can be achieved as:
\begin{equation}
F_R^L(i,k,c)= F_O^L(i,p(k),c),\quad k \in \{1,...,\epsilon w_O^L\},
\label{eq:ret_maps}
\end{equation}
where $p$ is a list that contains all the elements in $\mathcal{P}$ and is sorted from small to large.

Since deep CNN feature is extracted by non-linear mappings and cascade convolutions, previous retargeting methods based on minimum importance cost, such as seam carving and column removal, may destroy underlying relationship and distribution of CNN features, which causes structure distortion and/or content loss in the retargeted image (Figure \ref{fig:com_sc_cr_pcr}). The main advantage of UrS is that it avoids distortions (e.g., content loss, structure disortion) from columns/rows over-removing in regions of low feature importance but high structure importance, and meanwhile preserves the semantic importance of CNN features. We detail the comparison between UrS and other retargeting method when being applied in the DeepIR framework in Section \ref{comb_deepir}, which verifies the effectiveness of UrS.


\subsection{Reconstruction}
\begin{figure}[!htbp]
\centering
\includegraphics[width=8.0cm]{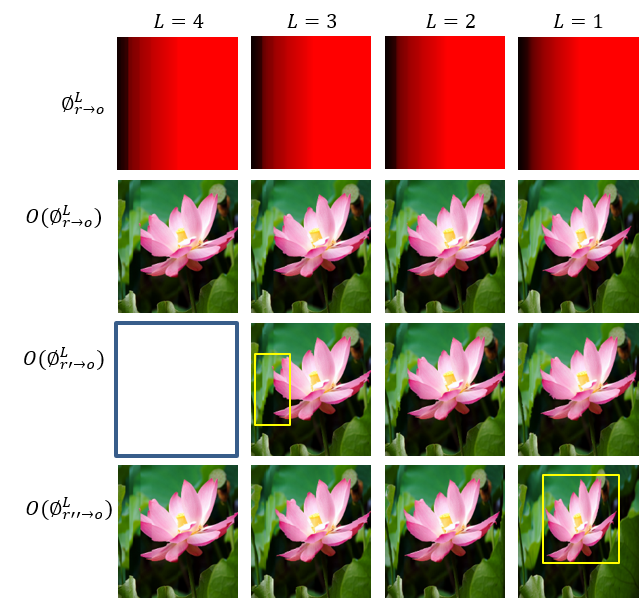}
	\caption{Visualization of coarse-to-fine reconstructed results with step-by-step NNF fusion. First row shows $\phi_{r\rightarrow o}^{L}$ in each layer. The following three rows show the retargeted images obtained by different NNF mapping functions.}
	\label{fig:NNF_reconst}
\end{figure}
After obtaining the retargeted image's feature maps at the highest layer (i.e., $F_R^4$), the following question is how to propagate the $F_R^4$ back to the lower layers. As presented in \cite{Liao:2017:VAT:3072959.3073683}, this problem can be solved by minimizing the following loss function:
\begin{equation}
\ell_{F_{R'}^{L-1}} = ||\text{CNN}_{L-1}^{L}(F_{R'}^{L-1})-F_R^{L}||,
\label{eq:deconv}
\end{equation}
where $F_{R'}^{L-1}$ is the reconstructed feature maps in the level $L-1$, $\text{CNN}_{L-1}^{L}(\cdot)$ is the $L$th CNN layer. This optimization problem for feature deconvolution can be solved by gradient descent (L-BFGS) \cite{zhu1997algorithm}.

In the following three steps, we propose to obtain more appropriate feature maps $F_R^{L-1}$ from consideration of both high-level and low-level information. First, we estimate a NNF mapping function $\phi_{r'\rightarrow o}^{L-1}$ that maps a point in feature map $F_{R'}^{L-1}$ to another in $F_O^{L-1}$. Second, we obtain the second NNF mapping function $\phi_{r''\rightarrow o}^{L-1}$ between $F_O^{L-1}$ and $F_{R''}^{L-1}$, where $F_{R''}^{L-1}$ is deep features resized from $F_O^{L-1}$ by UrS. Then we propose to fuse $\phi_{r'\rightarrow o}^{L-1}$ and $\phi_{r''\rightarrow o}^{L-1}$ using a weighted sum:
\begin{equation}
\phi_{r\rightarrow o}^{L-1} = \alpha^{L-1} \phi_{r'\rightarrow o}^{L-1} + (1-\alpha^{L-1})\phi_{r''\rightarrow o}^{L-1},
\label{eq:fusion}
\end{equation}
where $\alpha^{L-1}$ controls the trade-off between current layer's and previous layer's mapping information. Finally, the $F_R^{L-1}$ is obtained by warping $F_O^{L-1}$ with $\phi_{r\rightarrow o}^{L-1}$: $F_R^{L-1}=F_O^{L-1}(\phi_{r\rightarrow o}^{L-1})$. Figure \ref{fig:NNF_reconst} shows that our retargeted results are gradually updated from coarse to fine. We can observe that retargeted image $O(\phi_{r'\rightarrow o}^{L})$ constructed directly from $F_{R'}^{L}$ usually lacks the constraint of low-level details and suffers leaf unsmoothness in high layers (e.g., $O(\phi_{r'\rightarrow o}^{3})$). Though $O(\phi_{r''\rightarrow o}^{L})$ contains relatively smoother content, it lacks the supervision of high-level semantic information in low layer results (e.g., $O(\phi_{r''\rightarrow o}^{1})$). So $O(\phi_{r\rightarrow o}^{L})$ is the result from achieved balance between high-level semantic content and low-level details.

After we obtain NNF  $\phi_{r\rightarrow o}^1$ at the lowest feature layer, we assume that the pixel level mapping function $\Phi_{r\rightarrow o}$ is equal to $\phi_{r\rightarrow o}^1$.  Then we reconstruct retargeted image $R$: $R=\frac{1}{n}\sum_{x\in N(p)}(O(\Phi_{r\rightarrow o}(x)))$, where $n=5\times 5$ is the size of patch $N(p)$.

\begin{algorithm}
\caption{DeepIR Framework}
\label{alg_1}
\begin{algorithmic}[1]
\Require One RGB image $O$, target aspect ratio $\epsilon$.
\Ensure One pixel-location mapping function $\Phi_{r\rightarrow o}$ and one retargeted image $R$.
\State \textbf{Preprocessing}:
\State $\{F_O^L\} (L=1...4)$ $\gets$ feeding $O$ to VGG-19.
\State $F_{R''}^{4}$ $\gets$ resizing $F_{O}^{4}$ by UrS and $\epsilon$.
\State $F_{R}^{4} \gets F_{R''}^{4}$.
\State \textbf{Reconstruction}
\For{$L=4$ $to$ $2$}
\State $F_{R'}^{L-1}$ $\gets$ solving loss function Eqn.(\ref{eq:deconv}).
\State $F_{R''}^{L-1}$ $\gets$ resizing $F_{O}^{L-1}$ by UrS and $\epsilon$.
\State $\phi_{r'\rightarrow o}^{L-1}$ $\gets$ mapping $F_{R'}^{L-1}$ to $F_{O}^{L-1}$.
\State $\phi_{r''\rightarrow o}^{L-1}$ $\gets$ mapping $F_{R''}^{L-1}$ to $F_{O}^{L-1}$.
\State $\phi_{r\rightarrow o}^{L-1}$ $\gets$ fusing NNFs by Eqn.(\ref{eq:fusion}).
\State $F_{R}^{L-1}$ $\gets$ $F_{O}^{L-1}(\phi_{r\rightarrow o}^{L-1})$.
\EndFor
\State $\Phi_{r\rightarrow o}\gets\phi_{r\rightarrow o}^1$.
\State $R\gets\frac{1}{n}\sum_{x\in N(p)}(O(\Phi_{r\rightarrow o}(x)))$.

\end{algorithmic}
\end{algorithm}
\subsection{Algorithm and Performance}

The pseudo code of our implementation is shown in Algorithm \ref{alg_1}. We have implemented our algorithm and conducted experiments on one NVIDIA K80 GPU. The time of retargeting images in our experiments ranges from $20$ seconds to $60$ seconds, which depends on the sizes of input images. Specially, the time of PatchMatch ranges from $8$ seconds to $16$ seconds. The feature deconvolution may require $10$ seconds to $35$ seconds to converge.
\section{Experiment Results}\label{experiment}
\begin{figure}[!htb]
	\centering
     \includegraphics[width=0.95\linewidth]{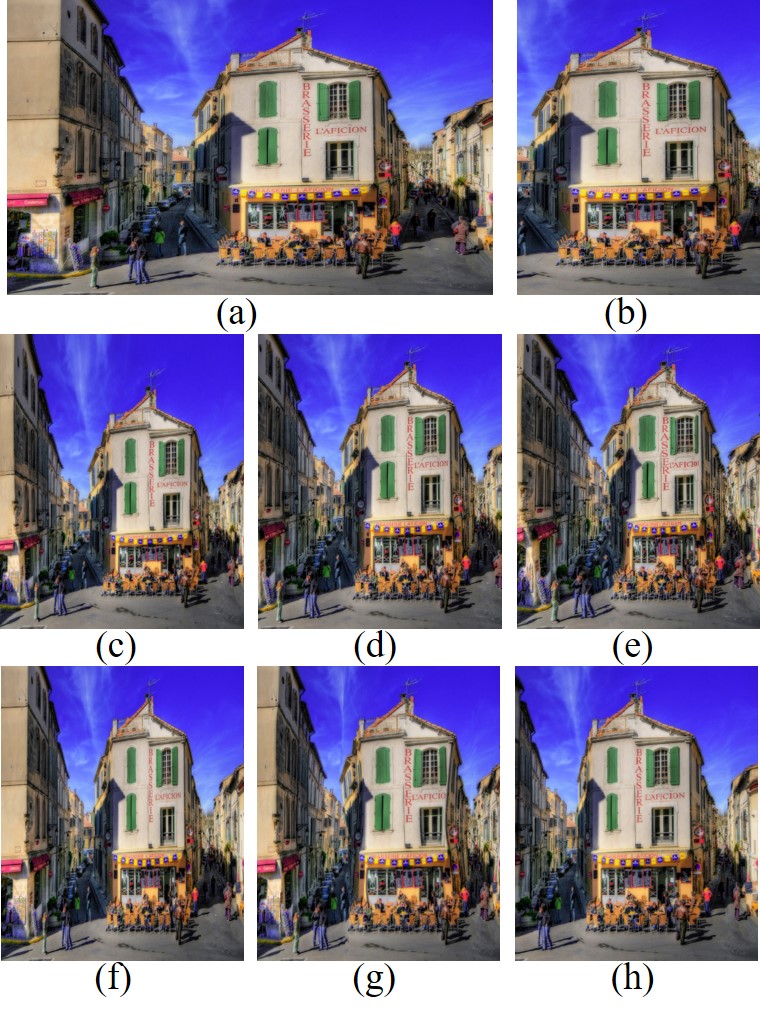}
	\caption{Comparison on ``Brasserie$\_$L$\_$Aficion'' in RetargeMe. (a) Original image. (b) CR. (c) SV. (d) MULTIOP. (e) SC. (f) SCL. (g) WARP. (h) Ours.}
	\label{fig:com_stoa_1}
\end{figure}
We conduct experiments on the widely used RetargetMe benchmark \cite{rubinstein2010comparative}. The RetargetMe dataset contains real-world images with various attributes and contents, so it's quite suitable to test the robustness and generality of retargeting method. We provide more experimental results in the supplementary document.

\subsection{Qualitative Evaluation}

\begin{figure}[!htb]
	\centering
\includegraphics[width=0.95\linewidth]{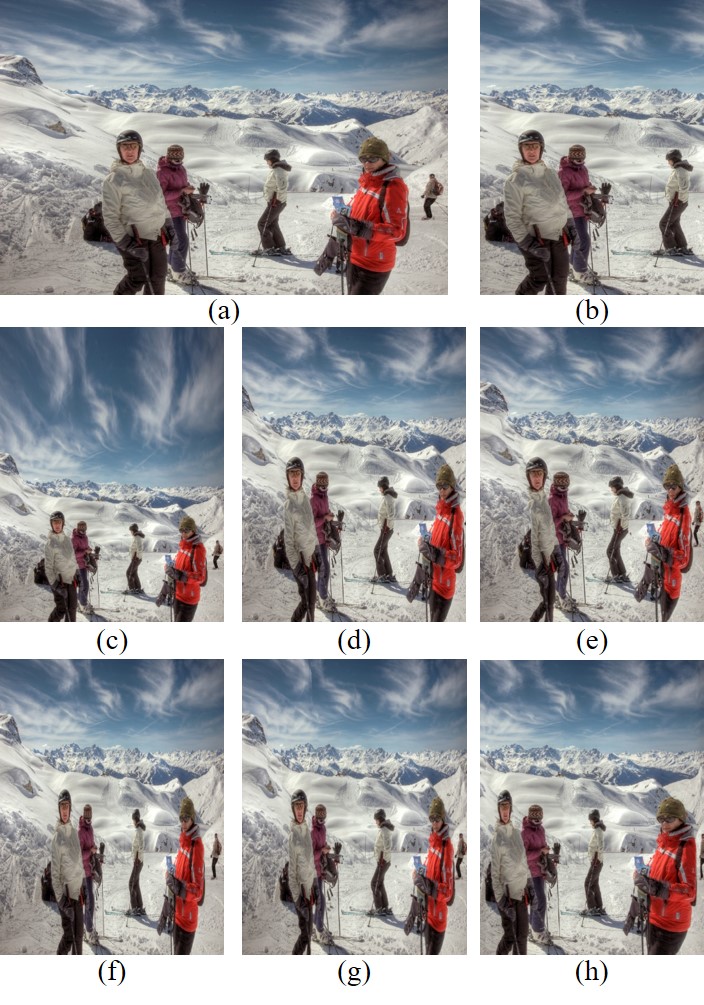}
	\caption{Comparison on ``ski'' in RetargeMe. (a) Original image. (b) CR. (c) SV. (d) MULTIOP. (e) SC. (f) SCL.  (g) WARP. (h) Ours.}
	\label{fig:com_stoa_4}
\end{figure}

Discrete and continuous image retargeting methods are compared in this section, such as manually chosen Cropping (CR), Streaming Video (SV) \cite{krahenbuhl2009system}, multi-operator (MULTIOP) \cite{rubinstein2009multi}, Seam Carving (SC) \cite{avidan2007seam}, uniform Scaling (SCL) and Nonhomogeneous warping (WARP) \cite{wolf2007non}. We compare our method with these algorithms on RetargetMe images as shown in Figure \ref{fig:com_stoa_1}, and Figure \ref{fig:com_stoa_4}. We can observe that: (1) Although CR tries to shrink image by removing parts of it, some content of semantic importance is lost, such as streets and buildings in Figure \ref{fig:com_stoa_1}(b), and skiers in Figure \ref{fig:com_stoa_4}(b). (2) SV can cause retargeted image to be out of proportion(i.e., some objects in retargeted image are over-shrunken or over-stretched compared to others) since  SV requires a fixed global scaling factor for the entire image, the examples are shown in Figure \ref{fig:com_stoa_1}(c). (3) SC and WARP tend to create distortions of lines and edges, and deformation of people and objects when there are not enough homogenous contents in the input image as shown in Figure \ref{fig:com_stoa_1}(e,g) and Figure \ref{fig:com_stoa_4}(e). (4)MULTIOP combines three operators (i.e., CR, SCL and SC) to avoid drawbacks of single operators. Although it outperforms single operators when the scenario is relatively simple (Figure \ref{fig:com_stoa_1}(d)), it can not preserve important contents as well as our method due to using low-level saliency detection, as shown in Figure \ref{fig:com_stoa_4}(d). (5) By using deep CNN for deep feature extraction, our method can effectively locate semantic contents as shown in Figure \ref{fig:com_stoa_1}(h) and Figure \ref{fig:com_stoa_4}(h). The results also demonstrate that our method can produce visually preferred results by using UrS and NNF fusion. Note that, although our method utilizes deep CNN for better semantic areas selection, our method does not need any training procedure given a pre-trained deep CNN. So such a framework we propose can be combined with other retargeting algorithms (feasibility is proved in Section \ref{comb_deepir}), which bridges the gap between traditional low level retargeting methods and newly-developed deep features.

\subsection{Quantitative Evaluation}
\begin{table}[htb!]
\centering
\begin{tabular}{p{1cm} p{1cm} p{1cm} p{1cm} p{1cm} p{1cm} p{1cm}}
\toprule
& \textbf{Ours} & \textbf{CR}& \textbf{SV}& \textbf{MUL-} \newline \textbf{TIOP}& \textbf{SCL} \\
\midrule
User Study & \textbf{101} & 36 & 98 & 90 & 35 \\
FRR   & \textbf{0.5307} & 0.5239 & 0.5091 & 0.5279 & 0.5215 \\
FD & 5.3349 & \textbf{2.8944} & 6.3555 & 5.7765 & 6.338 \\
\bottomrule
\end{tabular}
\caption{Objective and subjective scores of different methods. Best scores are in bold.}
\label{tab:userstudy}
\end{table}
\begin{figure*}[!htb]
	\centering
\includegraphics[width=0.8\linewidth]{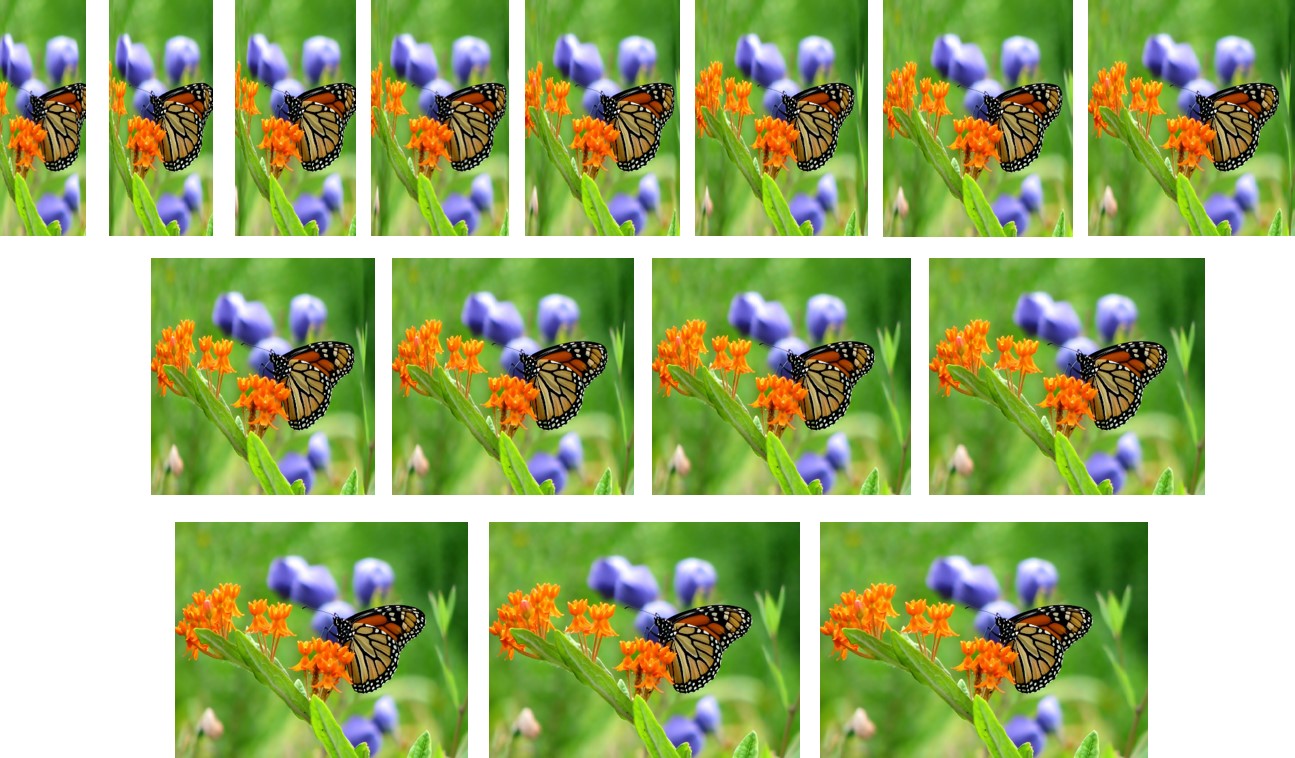}
	\caption{Influence of aspect ratio $\epsilon$ on retargeted results. From left to right and from top to down, the aspect ratio $\epsilon$ increases $0.5$ for each retageted image.}
	\label{fig:epsilon}
\end{figure*}
\begin{figure*}[!htb]
	\centering
\includegraphics[width=\linewidth]{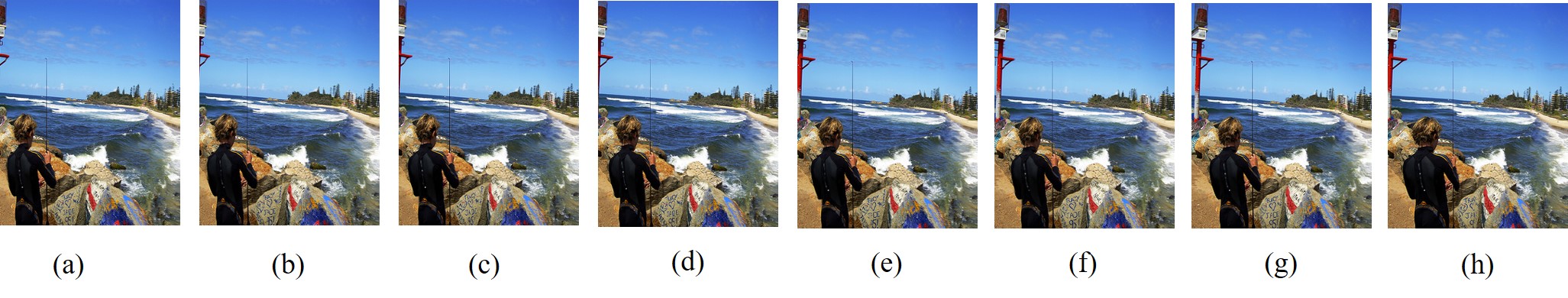}
	\caption{Influence of $\{\alpha^L\}_{L=1}^{3}$ on retargeted results. (a) $\{\alpha^L\}_{L=1}^{3}=\{0.1,0.2,0.3\}$. (b) $\{\alpha^L\}_{L=1}^{3}=\{0.2,0.3,0.4\}$. (c) $\{\alpha^L\}_{L=1}^{3}=\{0.3,0.4,0.5\}$. (d) $\{\alpha^L\}_{L=1}^{3}=\{0.4,0.5,0.6\}$. (e) $\{\alpha^L\}_{L=1}^{3}=\{0.5,0.6,0.7\}$. (f) $\{\alpha^L\}_{L=1}^{3}=\{0.6,0.7,0.8\}$. (g) $\{\alpha^L\}_{L=1}^{3}=\{0.7,0.8,0.9\}$. (h) $\{\alpha^L\}_{L=1}^{3}=\{0.8,0.9,1.0\}$.}
	\label{fig:alpha}
\end{figure*}
For quantitative evaluation, we compare with SCL and three state-of-the-art retargeting methods, i.e., CR, SV and MULTIOP, as reported in \cite{rubinstein2010comparative}. We first carry out a user study to assess quality of retageted images. We ask total $18$ subjects ($10$ males, $8$ females, age range $20-35$) from different backgrounds to make comparison of $20$ sets of retargeted images. The retargeted images are selected from RetargetMe of aspect ratio $\epsilon=0.5$, which is easier for subjects to judge. We show the subjects original image, our result and results from other methods. Then each subject selects the favorite retargeted images over other images. Then we also calculate two empirical objective scores: feature remain ratio (FRR), i.e.,
\begin{equation}
\text{FRR}=\frac{1}{4}\sum_{L=1}^{4}\frac{\sum_{i,j,c}F_R^L(i,j,c)}{\sum_{i,j,c}F_O^L(i,j,c)},
\label{eq:FRR}
\end{equation}
which measures the proportion of deep features remaining in the retargeted images, and feature dissimilarity (FD), i.e.,
\begin{equation}
\text{FD}=\frac{1}{4}\sum_{L=1}^{4}\sum_{i,j,c}\Vert F_O^L(\phi_{r\rightarrow o}^{L})(i,j,c)-F_R^L(i,j,c)\Vert^2,
\label{eq:FD}
\end{equation}
which calculates the square difference between original and retargeted images in the feature space. $\{F_R^L\}$ is obtained by feeding retargeted images to the VGG-19. The larger FRR and lower FD score is, the better image quality is. As shown in the Table \ref{tab:userstudy}, our method achieves best performance in terms of user study and FRR score, and second best performance in FD score. CR achieves the best FD score because it maintains original regions in the original images, which results in highly similar deep features as original ones'. Compared with SV which relies on both human labeled and automatic features, the quantitative results suggest the effectiveness of our method on semantic structure preserving.

\subsection{Robustness of DeepIR}
In order to investigate the influence of different aspect ratios $\epsilon$ and balancing parameters $\{\alpha^L\}_{L=1}^{3}$, we show the results of progressively increasing $\epsilon$ for retargeted image in Figure \ref{fig:epsilon} and results of varying $\{\alpha^L\}$ in Figure \ref{fig:alpha}. From Figure \ref{fig:epsilon}, we can observe that our method can effectively resize an image continuously while preserving important objects with various aspect ratios $\epsilon$. From Figure \ref{fig:alpha}, we can observe that the larger $\{\alpha^L\}$ tends to preserve more important contents, while too large $\{\alpha^L\}$ may lead to information loss, such as incomplete railing in the left of Figure \ref{fig:alpha}(h). The smaller $\{\alpha^L\}$ also has its drawback at preserving important content in image, such as the over-squeezed boy in Figure \ref{fig:alpha}(a). Therefore, $\{\alpha^L\}=\{0.7,0.8,0.9\}$ is chose as our configuration to achieve a trade-off between high-level semantic content and low-level details.

\subsection{Combining Retargeting Methods with DeepIR}\label{comb_deepir}
\begin{figure}[!htb]
	\centering
    \includegraphics[width=8.0cm]{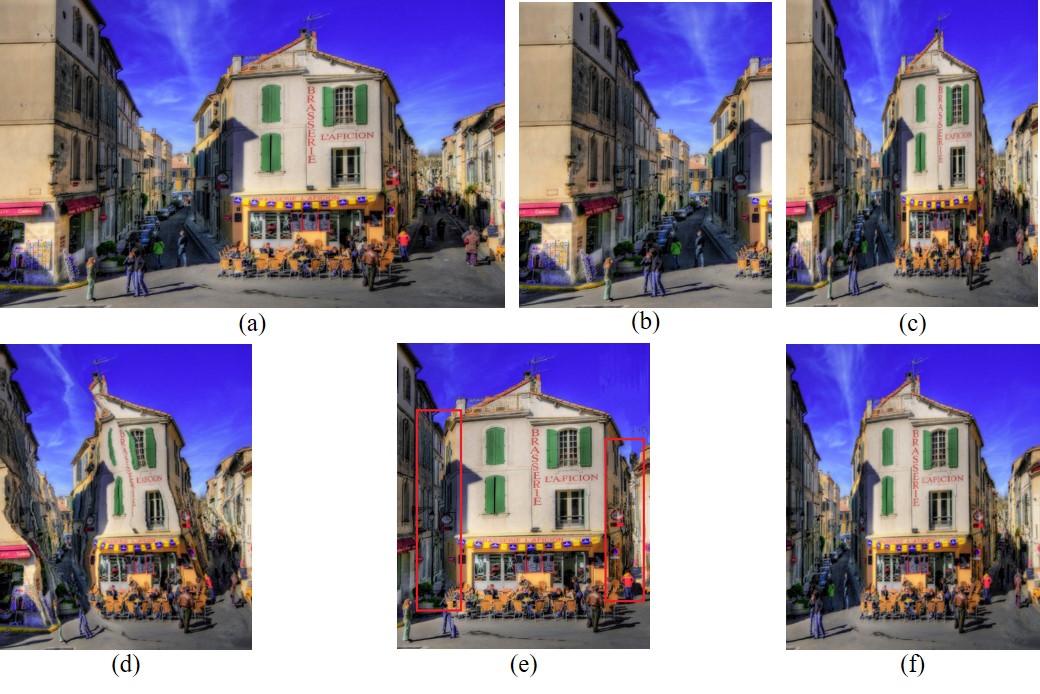}
	\caption{One example of using CR, SCL, SC, column removal and our UrS for deep feature retargeting in DeepIR framework. (a) Original image. (b) DeepIR with CR. (c) DeepIR with SCL. (d) DeepIR with SC suffers structure distortion. (e) DeepIR with column removal suffers content discontinuity. (f) DeepIR with UrS produces visually preferred result.}
	\label{fig:com_sc_cr_pcr}
\end{figure}
In order to verify the effectiveness of UrS on feature retargeting and flexibility of our DeepIR, we combine other retargeting methods with DeeIR framework, including CR, SCL, SC and column removal that removes columns/rows based on minimum importance cost. In each layer, the feature maps are resized by these retargeting methods instead of UrS and propogated to the lowest layer to generate final retargeted images. The results are illustrated in Figure \ref{fig:com_sc_cr_pcr}. However, we can observe that most of these methods cannot utilize the advantages of deep features and even not able to surpass the corresponding low-level retargeting method. Specifically, CR inevitably removes semantically important content in the original image (Figure \ref{fig:com_sc_cr_pcr}(b)). In Figure \ref{fig:com_sc_cr_pcr}(c), SCL simply stretches the object to the target aspect ratio as its behavior in Figure \ref{fig:com_stoa_1}(f). Also, SC removes the least importance seams in the feature space and may destroy the semantic structures in the deep features, which results in structure non-homogeneity in the retargeted image (Figure \ref{fig:com_sc_cr_pcr}(d)). Similar problem is also raised by column removal in Figure \ref{fig:com_sc_cr_pcr}(e), where the content of retargeted image is discontinuous. To summarize, the proposed UrS method that resizes feature maps based on cumulative obscurity map sampling can best maintain the semantic structure of original image and eliminate the discontinuity in retargeted image so far. Although previous retargeting methods based on saliency map are not suitable for deep feature retargeting, we could find that DeepIR is flexible to incorporate other image retargeting methods into the framework.
\subsection{Failure Cases}
\begin{figure}[!htb]
	\centering
\includegraphics[width=0.8\linewidth]{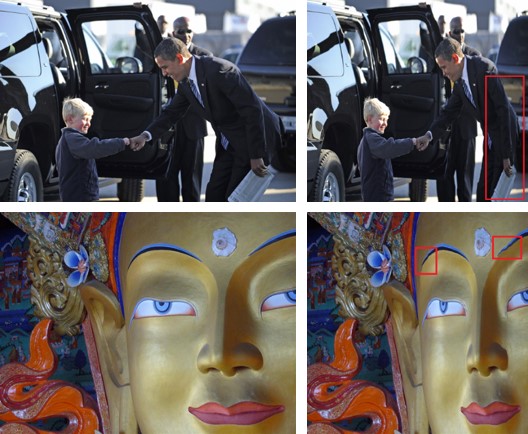}
	\caption{Failure cases. First row left: ``obama'' image. First row right: retargeted ``obama'' image, $\epsilon=0.75$. Second row left: ``buddha'' image. Second row right: retargeted ``buddha'' image, $\epsilon=0.75$.}
	\label{fig:failcases}
\end{figure}
The deep image retargeting technique relies on two previous works: the pre-trained VGG-19 network \cite{simonyan2014very} and PatchMatch \cite{barnes2009patchmatch}. Therefore, due to the limited capability of representing all objects and visual structures, we find that the DeepIR may over squeeze the important areas or over maintain the less important areas in some cases. We show an example in the first row of Figure \ref{fig:failcases}, where the body of Obama is over squeezed and the car is over preserved. We expect there will be a network with stronger representation ability in the future. For PatchMatch in the deep feature space, there still exists mismatching in some cases as shown in the second row of Figure \ref{fig:failcases}, where the eyebrow of Buddha suffers some discontinuous mismatching. One possible solution is to improve PatchMatch for scale-invariant matching in the future.
\section{Conclusions}\label{conclusion}
In this paper, we propose a new technique called \emph{Deep Image Retargeting} (\emph{DeepIR}). Our method utilizes deep CNN for the content-aware image retargeting. We first use a pre-trained deep convolutional neural network to extract deep features of original image. Then we propose a uniform re-sampling (UrS) image retargeting method to resize feature maps at each feature layer. The UrS can effectively preserve semantic structure as well as maintain content smoothness in the retargeted image. Finally, we reconstruct the deep features of retargeted image in each layer through a coarse-to-fine NNF search and a step-by-step NNF fusion. The retargeted image is obtained by using spatial correspondence at the lowest layer. We have carried out sufficient experiments to validate the effectiveness of our proposed technique on RetargetMe dataset. In general, our method solves the problem of applying pre-trained deep neural network for content-aware image retargeting. Our model is flexible and one can easily combine other retargeting algorithms with proposed framework for future works.

\clearpage
\bibliographystyle{aaai}
\bibliography{egbib}

\end{document}